\documentclass[letterpaper, 10 pt, journal, twoside]{IEEEtran}

\usepackage{graphicx} 
\usepackage{amsmath} 
\usepackage{amssymb}  
\usepackage{booktabs}
\usepackage{pgfplots}
\usepackage{hyperref}

\usepackage{enumitem}

\title{Fast and Robust Registration of Partially Overlapping Point Clouds}

\author{Eduardo~Arnold, Sajjad~Mozaffari and Mehrdad~Dianati
	\thanks{Manuscript received: September 9, 2021; Revised November 23, 2021; Accepted December 17, 2021.}
	\thanks{This paper was recommended for publication by Editor Sven Behnke upon evaluation of the Associate Editor and Reviewers' comments.}%
	\thanks{This work was supported by Jaguar Land Rover and the U.K.-EPSRC as part of the jointly funded Towards Autonomy: Smart and Connected Control (TASCC) Programme under Grant EP/N01300X/1.}
	\thanks{All authors are with the WMG, University of Warwick, Coventry, U.K. {\tt\footnotesize Email: e.arnold@warwick.ac.uk}}%
	\thanks{Digital Object Identifier (DOI): see top of this page.}
}

\newcommand{\R}{\mathbb{R}}
\newcommand{\T}{\mathsf{T}}

\DeclareMathOperator*{\argmax}{arg\,max}
\DeclareMathOperator{\Tr}{Tr}

\begin{document}
	
	\maketitle
	
	\begin{abstract}
Real-time registration of partially overlapping point clouds has emerging applications in cooperative perception for autonomous vehicles and multi-agent SLAM.
The relative translation between point clouds in these applications is higher than in traditional SLAM and odometry applications, which challenges the identification of correspondences and a successful registration.
In this paper, we propose a novel registration method for partially overlapping point clouds where correspondences are learned using an efficient point-wise feature encoder, and refined using a graph-based attention network.
This attention network exploits geometrical relationships between key points to improve the matching in point clouds with low overlap.
At inference time, the relative pose transformation is obtained by robustly fitting the correspondences through sample consensus.
The evaluation is performed on the KITTI dataset and a novel synthetic dataset including low-overlapping point clouds with displacements of up to 30m.
The proposed method achieves on-par performance with state-of-the-art methods on the KITTI dataset, and outperforms existing methods for low overlapping point clouds.
Additionally, the proposed method achieves significantly faster inference times, as low as 410ms, between 5 and 35 times faster than competing methods.
Our code and dataset are available at \url{https://github.com/eduardohenriquearnold/fastreg}.
\end{abstract}


\section{\uppercase{Introduction}}
\label{sec:introduction}
	\IEEEPARstart{P}{oint} cloud registration is the problem of estimating the rigid relative pose transformation that aligns a pair of point clouds into the same coordinate system.
	This is a key problem in many downstream applications including 3D scene reconstruction \cite{li20183d}, localisation \cite{Lu_2019_CVPR} and SLAM \cite{ramezani2020online}.
	Recent applications such as Augmented Reality (AR) \cite{gao2017stable}, cooperative (multi-agent) perception for autonomous vehicles \cite{arnold2020cooperative} and multi-agent SLAM \cite{dube2017online} introduce new challenges to this problem.
	Specifically, these applications require registration methods that are robust to point clouds with low overlap, \textit{e.g.} when sensors are far apart, and capable of operating in real-time.
	
	\begin{figure}
		\centering
		\input{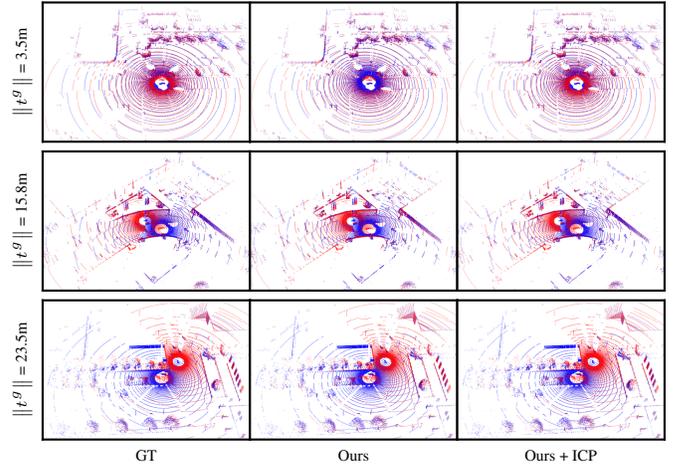}
		\caption{Qualitative results of the proposed method. Each row represents a different sample from the CODD test set and the vertical label indicates the relative translation between point clouds in meters, where $t^g$ indicates the ground-truth relative translation vector.}
		\label{fig:qualitative}
	\end{figure}
	
	Existing registration methods are often designed and evaluated assuming a significant overlap between the input point clouds.
	This assumption is valid for applications such as SLAM \cite{ramezani2020online} and lidar odometry \cite{graeter2018limo}, where pairs of point clouds are obtained sequentially in adjacent time steps by a single vehicle navigating in a driving environment.
	On the other hand, applications such as cooperative perception \cite{arnold2020cooperative} and multi-agent SLAM \cite{dube2017online} require registering point clouds obtained simultaneously from a pair of sensors on two different vehicles that are potentially far apart, and thus, may have low field-of-view overlap, \textit{e.g.} Figure \ref{fig:qualitative}.
	As the relative translation between the sensors increases, the number of identifiable correspondences decreases, which poses challenges in registering the point clouds accurately.
	
	The majority of existing point cloud registration methods cannot guarantee real-time execution.
	Traditional local registration methods such as Iterative Closest Point (ICP) \cite{arun1987least} solve the problem iteratively assuming an initial relative pose.
	However, the iterative nature of such methods renders them unfeasible for real-time applications, particularly considering large scale point clouds.
	These methods are also prone to non-optimal solutions when the initial pose estimate is poor, which may be addressed with global-optimisation variants \cite{olsson2008branch,yang2013go} at the cost of higher computational complexity.
	Another category of methods identify correspondences between point clouds using a distance metric between hand-engineered features \cite{rusu2009fpfh} or learned point-wise features \cite{choy2019fully}.
	These correspondences are often contaminated by a large number outliers and must be filtered using RANSAC \cite{fischler1981random,bustos2017guaranteed} or learned models \cite{choy2020deep}, which further increases the registration execution time.
	Furthermore, state-of-the-art learning-based models \cite{choy2019fully,choy2020deep} require computationally demanding 3D convolutions and generate numerous putative correspondences, introducing a bottleneck on the RANSAC loop and rendering real-time execution unfeasible.

	To mitigate the aforementioned limitations, we propose a novel point cloud registration method capable of operating in real-time and robust to low-overlapping point clouds.
	The proposed method identifies correspondences between the source and target point clouds by learning point-wise features.
	A novel encoder hierarchically subsamples the point clouds to reduce the number of key points and improve the run-time performance.
	The resulting features are refined using self- and cross-attention based on a graph neural network.
	The attention network leverages geometrical relationships between key points and their features across to improve the correspondence accuracy, particularly in regions of low overlap.
	The relative pose parameters are obtained by fitting the learned correspondences using Random Sample Consensus (RANSAC) to robustly reject outliers.	
	During inference, the RANSAC fitting is done efficiently considering a small number of correspondences, which allows end-to-end inference times below 410ms.
	The model is trained and evaluated separately on the KITTI odometry dataset and a novel Cooperative Driving Dataset (CODD).
	The relative translation between sensors in CODD ranges up to 30m, introducing challenging pairs of point clouds with low overlap, which we hope will create a new research benchmark.
	Our contributions are summarised as:
	\begin{itemize}
		\item A computationally efficient point-wise feature encoder that allows identifying correspondences between point clouds;
		\item A graph neural network that provides self- and cross-attention between point clouds and improves the quality of correspondences;
		\item A novel registration method for point clouds that is robust to partially-overlapping point clouds and capable of operating in real-time;
		\item A new synthetic lidar dataset containing low overlapping point clouds in a wide range of driving scenarios;
	\end{itemize}

\section{\uppercase{Related Works}}
\label{sec:relatedwork}
	This section reviews existing point cloud registration methods in the literature and highlights how the method proposed in this paper differs from these existing works.
	We divide existing methods in the literature into two categories: traditional registration methods and learning-based methods.

	\subsection{Traditional Registration Methods}
		Iterative Closest Point (ICP) \cite{arun1987least} is a local registration method that assumes an initial relative pose and iteratively computes the transformation parameters that minimise the distance between each point in the source point cloud and its closest neighbour in the target point cloud.
		This method is highly sensitive to the initial pose estimate, and converges to non-optimal local-minima results when the initial pose estimate is poor.
		To mitigate this, \cite{olsson2008branch,yang2013go} estimate global optimum solutions for ICP considering branch-and-bound search over the transformation space.
		However, such global methods have significantly higher execution times, which prevents their usage in real-time applications.
				
		Other traditional approaches use handcrafted features to find correspondences between the point clouds.
		Fast Point Feature Histogram (FPFH) \cite{rusu2009fpfh} encodes the local geometry of 3D points using multi-dimensional feature vectors.
		But the correspondences obtained by comparing FPFH features are often contaminated by a large number of outliers, which prevents accurate registration.
		For this reason, RANSAC methods are used to filter out the outlier correspondences.
		More recently, TEASER \cite{Yang20tro-teaser} reformulates the registration problem using a truncated least-squares cost, which results in improved registration accuracy compared to RANSAC when considering a high number of outliers.

	\subsection{Learning-based Methods}
		One group of learning-based registration methods focus on learning accurate correspondences between the point clouds.
		Generally, these methods learn a mapping from the original Euclidean space to a latent feature space and optimise the mapping such that corresponding points have a small distance in the latent space.
		Deng \textit{et al.} \cite{deng2018ppfnet} uses a PointNet \cite{qi2017pointnet} model to learn point-wise features and trains the model using an $N$-tuple loss.
		VCR-Net \cite{wei2020endtoend} learns point-wise feature vectors using multi-layer-perceptrons (MLPs) to extract local features, which are refined using global attention and used to identify correspondences between point clouds.
		In contrast, \cite{choy2019fully} uses sparse fully convolutional networks to obtain voxel-wise features and trains the model using variations of triplet loss with hard negative mining.
		The resulting correspondences are often contaminated with outliers and need to be pruned using RANSAC \cite{fischler1981random} or further learning-based filtering \cite{choy2020deep} methods before estimating the pose transformation parameters.
		
		Another group of methods solve the problem end-to-end by learning the relative pose transformation directly.
		For example, PCRNet \cite{vsarode2019pcrnet} uses a PointNet \cite{qi2017pointnet} model to encode a global feature vector for both source and target point clouds and directly regress the transformation parameters.
		DeepVCP \cite{lu2019deepvcp} uses PointNet++ \cite{qi2017pointnetpp} to create point-wise features, then selects top K salient points using a learned weighting network to generate a deep feature embedding.
		The embeddings are fed to a 3D CNN to obtain soft-correspondences which are finally used to compute the transformation parameters in closed form.
		
		The proposed method differs from previous works in the following ways.
		First, our method considers a novel and computationally efficient point-wise feature encoder based on Set Abstraction (SA) and Feature Propagation (FP) layers \cite{qi2017pointnetpp}.
		While previous works \cite{lu2019deepvcp} have used PointNet++ feature encoders, we distinguish our encoder by adopting an architecture that hierarchically subsamples points at each layer, resulting in improved computational performance.
		Secondly, we improve the quality of the correspondences using a novel graph-based attention network that allows to efficiently combine self- and cross- information across point clouds.
		Differently from the feature-based attention in \cite{wei2020endtoend}, the proposed graph attention leverages both spatial and feature dimensions of local neighbourhoods to refine point-wise features.
		Finally, we extend our analysis beyond existing datasets and evaluate our model performance in challenging low overlapping point clouds using a novel dataset where the translation between sensor poses vary uniformly up to 30 meters.

\section{\uppercase{Problem Formulation}}
\label{sec:problem}
	Given two input point clouds $P_X  \subset \R^3$ and $P_Y \subset \R^3$, the registration problem is to estimate the rigid relative pose transformation that aligns $P_X$ into the coordinate system of $P_Y$.
	This transformation is parametrised by a rotation matrix $R \in SO(3)$ and a translation vector $t \in \R^3$.
	The problem can be solved by identifying pairs of correspondences between $P_X$ and $P_Y$.
	Given a set of correspondences, $X = \{x_1,\dots,x_N \} \subset P_X, Y = \{y_1,\dots,y_N \} \subset P_Y$, where $(x_i, y_i), i=1,\dots,N$ are correspondence pairs, the transformation parameters are obtained by the minimisation of the least-squares error:
	\begin{equation}
		\label{eq:ls-error}
		E(R,t) = \frac{1}{N} \sum_{i=1}^N \left\| Rx_i + t - y_i \right\|^2.
	\end{equation}
	This error is a form of the Orthogonal Procrustes problem \cite{1996weightedProcrustes} and admits the closed-form solution described below.
	First, the centroids are computed as
	\begin{equation}
		\bar{x} = \frac{1}{N}\sum_{i=1}^{N} x_i,\quad\quad  \bar{y} = \frac{1}{N}\sum_{i=1}^{N} y_i,
	\end{equation}
	and the covariance matrix, denoted by $H$, is obtained using
	\begin{equation}
		H = \sum_{i=1}^{N} (x_i - \bar{x})(y_i - \bar{y})^\T.
	\end{equation}
	Finally, the rotation matrix and translation vector $R,t$ that minimise Eq. \ref{eq:ls-error} are computed in closed-form as
	\begin{align}
		\label{eq:procrustes-closed-form}
		R &= V \left[\begin{smallmatrix} 1 & & \\ & 1 &  \\ & & \det(V^\T U) \end{smallmatrix}\right] U^\T, \\
		t &= -R\bar{x} + \bar{y}, \nonumber
	\end{align}
	considering the Singular Value Decomposition (SVD) $H = U S V^T$.
	The next section proposes a novel method to efficiently obtain correspondences between pairs of point clouds.
	
\section{\uppercase{Proposed Method}}
\label{sec:method}
\begin{figure*}
		\centering
		\includegraphics[width=\linewidth]{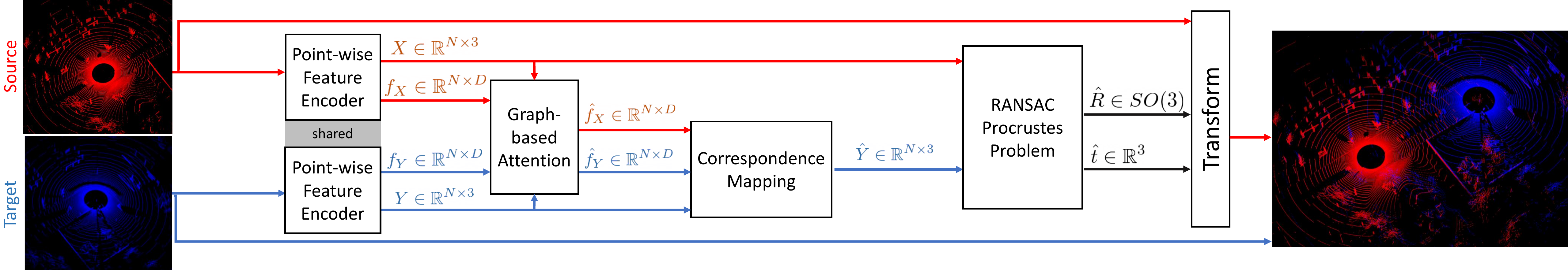}
		\caption{Pipeline and data flow of the proposed point cloud registration method.}
		\label{fig:diagram}
	\end{figure*}

	This section presents a novel method for robust point cloud registration using learned correspondences targetting efficient, real-time inference.
	Figure \ref{fig:diagram} describes the components and data flow of the proposed method.
	The proposed method can be summarised as follows:
	\begin{enumerate}[label=(\Alph*)]
		\item Both point clouds are fed to an encoder to obtain a subset of key points and associated point-wise features.
		\item A graph neural network refines the point-wise features considering self- and cross-attention.
		\item The resulting features are used to identify correspondences between the source and target key points.
		\item The relative transformation parameters $R,t$ are robustly estimated using a RANSAC formulation of the problem defined in Section \ref{sec:problem}. 
	\end{enumerate}
	The aforementioned components and the training process are described in the following subsections.
	
	\subsection{Point-wise Feature Encoder}
\begin{figure}
			\centering
			\includegraphics[width=\linewidth]{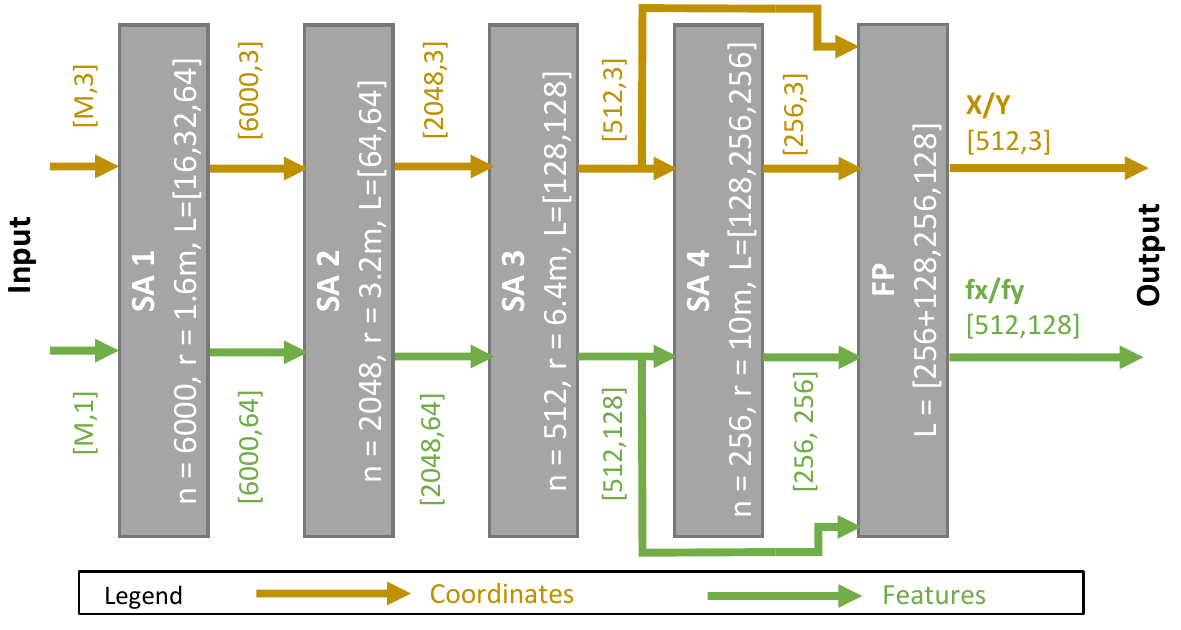}
			\caption{Point-wise feature encoder model architecture. The model consists of four Set-Abstraction (SA) layers and one Feature Propagation (FP) layer. Point coordinates and features are represented in yellow and green, respectively. The brackets indicate the dimensionality of each matrix. The number of input points, denoted by $M$, is arbitrary, and the model consistently outputs $N=512$ point coordinates and their respective feature vectors with $D=128$ dimensions.}
			\label{fig:encoder}
		\end{figure}

		The encoder is a core component of the pipeline, as it computes point-wise features that will be used to identify correspondences.
		We propose a novel and computationally efficient encoder network based on Set Abstraction (SA) and Feature Propagation (FP) layers \cite{qi2017pointnetpp}.
		The proposed encoder architecture, including the hyper-parameters of each layer, is depicted in Figure \ref{fig:encoder}.
		The encoder outputs subset of sampled coordinates (key points) from the source and target point clouds, denoted by $X$ and $Y$, and their respective feature vectors, $f_X$ and $f_Y$.
		The input to the encoder consists of 3D point coordinates and corresponding features, \textit{e.g.} lidar return intensity.
		Note that the input features are optional, but in this work they consist of a single scalar per point representing the lidar intensity.
		The source and target point clouds are fed to the encoder independently.
		
		The first four encoder layers are SA layers.
		A SA layer consists of four operations:
		\begin{enumerate}
			\item $n$ coordinates are sampled from the previous layer using Farthest Point Sampling (FPS) \cite{moenning2003fps}.
			\item A local neighbourhood of each sampled coordinate is established by selecting all points within radius $r$ of the respective coordinate.
			\item The features of the points in each neighbourhood are fed to a shared Multi Layer Perceptron (MLP), denoted as a list $L$ containing the number of intermediate nodes per layer.
			\item The resulting feature vectors of the $n$ sampled coordinates is computed using an aggregation function (\textit{max-pooling}) over the MLP output of the points in the respective neighbourhoods.
		\end{enumerate}
		Each SA layer hierarchically subsamples and aggregates information from the previous layer with progressively larger receptive volumes, which is a fundamental step in reducing the computational cost of our pipeline.
		At the same time, it is also important not to discard valuable information, \textit{i.e.} prioritising that points from one layer are within the neighbourhood of sampled points in the next layer.
		This trade-off is achieved by tuning the layers' hyper-parameters, namely $n,r,L$, such that the sampled points' neighbourhood include most points from the previous layer.
		These hyper-parameters were optimised for large outdoor driving environments and would require fine-tuning for indoor scenarios.
		
		The last encoder layer is an FP layer.
		It propagates high level information from SA4 to the points in the previous layer (SA3) as illustrated in Figure \ref{fig:encoder}.
		This is achieved by interpolating the feature vectors in SA3 layer using the features from the three nearest-neighbour coordinates in SA4.
		The final features are obtained fusing the original SA3 features with the interpolated SA4 features using a shared MLP, represented by a list $L$ of intermediate nodes.
		More details about the interpolation can be found in \cite{qi2017pointnetpp}.
			
	\subsection{Graph-based Attention}
\begin{figure}
			\centering
			\includegraphics[width=\linewidth]{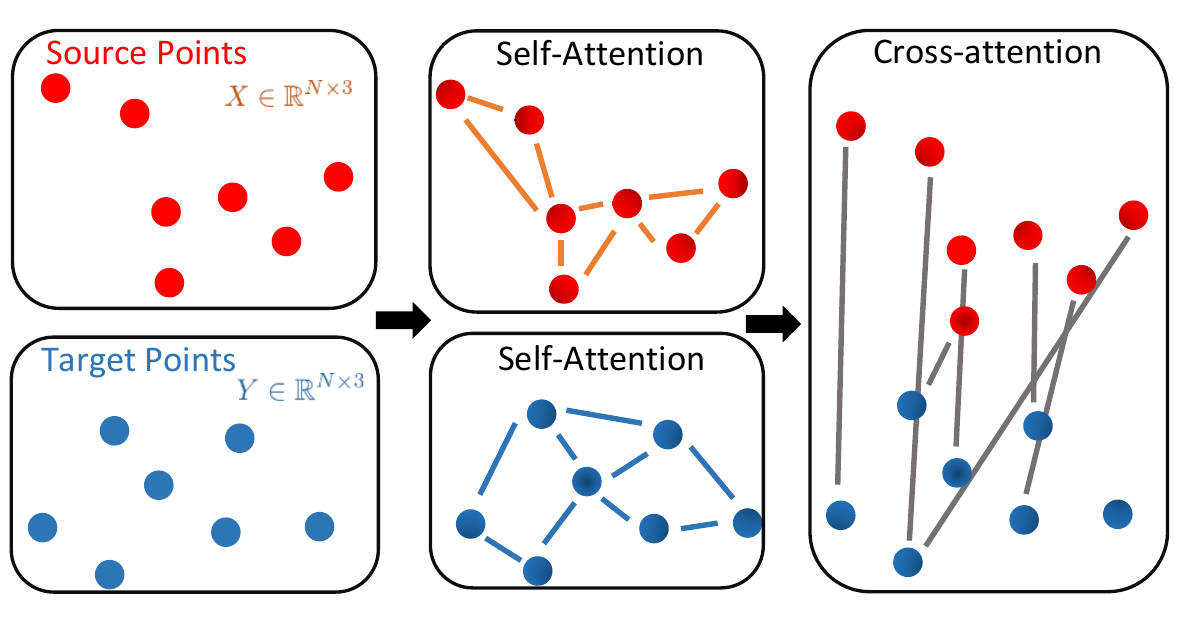}
			\caption{Graph-based attention representation. Points are represented as graph nodes. The nodes are defined by their position, \textit{i.e.} 3D coordinates, and their feature vector (not represented in the image). The graphs connecting the points are created based on the $k$-NN ($k=32$) between the points. In the self-attention layer, the $k$-NN distance is the Euclidean distance between points' spatial coordinates. In the cross-Attention graph, the $k$-NN distance is the dot product between points' feature vectors.}
			\label{fig:graphatt}
		\end{figure}

		The feature vectors obtained with the encoder network represent local point cloud information.
		However, these features are agnostic to the global context of the point cloud.
		For example, if a point cloud contains multiple objects, \textit{e.g.} trees, it would be difficult to distinguish between individual trees.
		Another problem, most critical for low overlapping point clouds, is that regions of overlap generally have different point densities in each point cloud, challenging the correct identification of correspondences since a point's features change with the density of points in its neighbourhood.
		To mitigate both problems, we propose a graph-based attention module which transforms points' feature vectors considering the wider point cloud context (self-attention) and the context of both source and target point-clouds (cross-attention).
		Self-attention increases the distinctiveness of key points by attending to their surrounding context.
		The cross-attention layer learns to refine point-wise features by attending to the most similar features across point clouds. 
		Both layers, illustrated in Figure \ref{fig:graphatt}, increase the likelihood of finding accurate correspondences, even in cases of low overlap.
		
		The self-attention layer introduces attention between points within the same point cloud.
		A graph connecting the points (nodes) is created for each point cloud using the $k$-Nearest-Neighbours of the points' spatial coordinates.
		Let $f_i$ be a feature vector from either $f_X$ or $f_Y$. 
		The self-attention layer computes a residual term for $f_i$ using the Crystal Graph Convolution Operation \cite{xie2018cgconv}:
		\begin{equation}
			\label{eq:cgconv}
			\hat{f}_i = f_i + \max_{j \in \mathcal{N}(i)} \sigma(z_{i,j} W_f) \odot \text{Softplus}(z_{i,j} W_s),
		\end{equation}
		where $\mathcal{N}(i)$ indicates the $k$ neighbour nodes of $i$, $z_{i,j} = [f_i, f_j] $ is the aggregated features of nodes $i,j$.
		$\sigma(\cdot)$ indicates the sigmoid function, and the \textit{Softplus} function is defined as $\text{Softplus}(z) = \log(1+e^z)$.
		The attention matrices $W_f, W_s$ are parameters to be learned and the operation $\odot$ represents element-wise multiplication.
		We adopt the number of nearest neighbours $k=32$, which provides a good trade-off between accuracy and computational efficiency.
		This process is performed independently with shared parameters for both source and target point clouds.
		
		Following the self-attention layer, the cross-attention layer allows interaction between the source and target point cloud features.
		This layer creates a bi-partite graph between source and target points.
		Each source point is connected to the $k$-Nearest-Neighbours nodes in the target point cloud, where the distance metric is the dot product between the feature vectors of the respective points.
		This layer uses the same residual update rule from Eq. \ref{eq:cgconv}, considering the different underlying graph and independent attention matrices $W_f', W_s'$.
		The graph-attention network output is given by the updated feature vectors from source and target point clouds, denoted by $\hat{f}_X$ and $\hat{f}_Y$, respectively.
		
	\subsection{Identifying Correspondences}
	\label{sec:soft-corr}
		The correspondences between the point clouds can be obtained by comparing the point-wise features of the source and target key points, denoted respectively as $\hat{f}_X, \hat{f}_Y \in \R^{N \times D}$.
		These feature vectors are normalised to unity $D$-dimensional vectors using the Euclidean norm.
		A matching probability map indicating the probability of correspondences between $X$ and $Y$ is computed as
		\begin{equation}
			\label{eq:softmap}
			\phi = \text{Softmax}\left( \frac{\hat{f}_X \cdot \hat{f}_Y^\T}{T} \right) \in \R^{N \times N},
		\end{equation}
		where $T$ is a temperature hyper-parameter and the \textit{Softmax} function is applied row-wise.
		Each element $\phi_{ij}$ represents the probability that the $i$-th key point in $X$ matches the $j$-th key point in $Y$.
		The \textit{Softmax} function scales the coefficients of each row $\phi_i$, ensuring a probability distribution over the points in $Y$.
		The temperature parameter, denoted by $T$, controls the entropy of distribution across points in $Y$.
		In the limit, when $T \to 0^+$, the coefficients become the one-hot encoding of the point in $Y$ with the highest similarity (\textit{i.e.} dot product).
		Finally, each key point $x_i \in X$ is matched to the key point in $Y$ with highest correspondence probability, resulting in the set of correspondence pairs $\{ (x_i, y_{\argmax_j \phi_{ij}}), i=1,\dots,N \}$.
		The ordered set of correspondences in $Y$ is denoted as $\hat{Y} = \{ y_{\argmax_j \phi_{1j}}, \dots, y_{\argmax_j \phi_{Nj}} \}$.	
		
	\subsection{Estimating Transformation Parameters}
		The previous step computes a correspondence for every point in $X$.
		In practice, only a fraction of points in $X$ will have correspondences in $Y$, particularly in the case of partially overlapping point clouds.
		Sensor noise and varying point densities can also lead to encoding errors and erroneous correspondences.
		To mitigate the effect of correspondence outliers, a common practice is to use sample consensus algorithms such as RANSAC \cite{fischler1981random,rusu2009fpfh}.
		A general version of this algorithm applied to this problem consists of three steps:
		\begin{enumerate}
			\item Create a hypothesis: Sample a minimal set of three correspondences from the set of correspondences and compute the transformation parameters using Eq. \ref{eq:procrustes-closed-form}.
			\item Score the hypothesis based on consensus: Compute the number of inlier correspondences, where a correspondence $(x_i,\hat{y}_i), x_i \in X, \hat{y}_i \in \hat{Y}$ is an inlier if $\left\| Rx_i + t - \hat{y}_i \right\| \leq \kappa$, where $\kappa$ is the inlier threshold, $R, t$ are the hypothesis parameters computed in the first step.
			\item Repeat the previous steps for $L$ times and select the hypothesis with the highest number of inliers.
		\end{enumerate}
		The number of tested hypotheses, denoted by $H$, offers a trade-off between computational performance and robustness to outliers.
		$H$ can also be derived to achieve a desired confidence of the selected hypothesis \cite{fischler1981random}, \textit{i.e.} sampling an outlier-free set of correspondences.
		While previous works used RANSAC with a large set of putative correspondences \cite{rusu2009fpfh}, this may be unfeasible for real-time systems.
		In this work, we achieve negligible RANSAC computational cost by learning a small set of correspondences ($N=512$), which allows to reduce the number of hypotheses being tested.
		
	\subsection{Training Process}
		The training process consists of optimising the encoder and attention networks to find accurate correspondences where they exist.
		We maximise the matching probabilities of ground-truth correspondences and minimise the matching probability of non-corresponding points.
		This is achieved by directly minimising the following loss function:
		\begin{equation}
			\label{eq:loss}
			\mathcal{L} = \frac{1}{N_c} \sum_{i=1}^N \delta_i \left[ -\phi_{i\hat{j}} + \frac{\lambda}{N-1} \sum_{j=1,j \neq \hat{j}}^N \phi_{ij} \right],
		\end{equation}
		where the binary variable $\delta_i$ indicates whether the source point $x_i$ has a correspondence in $Y$ and $\hat{j}$ represents the index of the corresponding point in $Y$. 
		Additionally, $N_c=\sum_{i=1}^N \delta_i$ is the number of ground-truth correspondences and $\lambda$ is a hyper-parameter scaling the contributions of incorrect matches into the loss function.
		A point in $X$ is considered to have a correspondence in $Y$ if, under the ground-truth transformation, it is within a distance smaller or equal to $1.6$ meters from a point in $Y$.
		This inlier distance is arbitrary and was chosen based on the smallest encoder radius.
		Data augmentation is employed by applying random rotation transformations to both input point clouds and adjusting the ground-truth rotation matrix accordingly.
		The optimisation details are described in Section \ref{sec:exp:implementation}.
		
\section{\uppercase{Performance Evaluation}}
\label{sec:exp}
	In this section, we first describe the datasets and the evaluation metrics, followed by the implementation details.
	We then compare the performance of the proposed method with traditional baselines, including ICP \cite{arun1987least}, FPFH RANSAC \cite{rusu2009fpfh} and TEASER \cite{Yang20tro-teaser}; and two state-of-the-art learning-based methods: FCGF \cite{choy2019fully} and DGR \cite{choy2020deep}.
	We also provide an ablation study identifying the impact of the proposed attention network into the registration performance.

	\subsection{Dataset}
	\label{sec:exp:dataset}
		The KITTI Odometry dataset \cite{Geiger2012CVPR} is traditionally used to evaluate point cloud registration methods in outdoor environments.
		We follow the evaluation protocol of recent methods \cite{FGR2016,choy2019fully,choy2020deep,Bai_2020_CVPR}, which adopt sequences 0 to 5 for training, 6 to 8 for validation and 9 to 10 for testing.
		In each sequence, the samples are created by selecting pairs of points clouds obtained sequentially by a single vehicle such that the translation between the poses is less than 10m.
		The ground-truth pose is provided by GPS and refined using ICP to reduce misalignment.
		
\begin{figure*}
			\centering
			\input{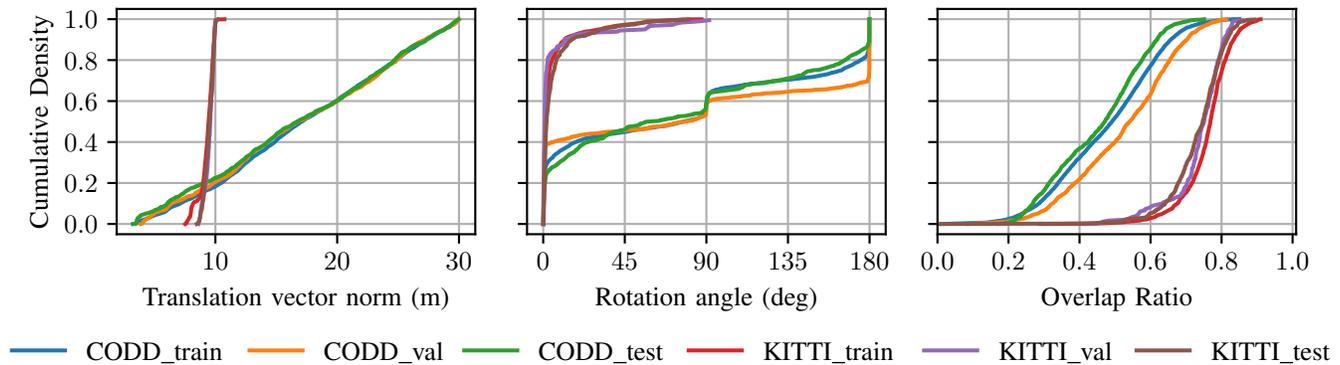}
			\caption{Cumulative density of the relative distance and rotation angle between coordinate systems of the pairs of point clouds in each subset.}
			\label{fig:plot:DatasetStats}
		\end{figure*}

		The distribution of poses in the KITTI dataset is limited to the trajectory of a single vehicle as it navigates the environment.
		In practice, registration methods must be resilient to point clouds with arbitrary relative pose, where the overlap between point clouds may vary significantly across samples.
		To this end, we introduce Cooperative Driving Dataset (CODD) \cite{codd}, an open-source synthetic dataset containing lidar point clouds collected simultaneously from multiple vehicles.
		This dataset is created using CARLA \cite{Dosovitskiy17} and features a diverse range of driving environments, including rural areas, suburbs, and dense urban centres.
		The dataset consists of 108 sequences, which are split into three independent subsets for training, validation and testing, as detailed in Table \ref{tab:dataset}.
		The samples are created by selecting all pair-wise combinations of point clouds obtained from vehicles driving simultaneously within a vicinity considering a maximum distance of 30m.
		Figure \ref{fig:plot:DatasetStats} presents the cumulative density plots of the relative distance (translation vector norm), rotation angle and overlap ratio of the pairs of point clouds in each dataset.
		The overlap ratio measures the overlap between point clouds as the percentage of points in the source point cloud that, when aligned, are within a distance smaller than $\gamma$ to any point in the target point cloud.
		Our CODD dataset has a significantly broader distribution of relative distance, rotation angles and overlap ratio between the point cloud pairs, which provides representative scenarios for cooperative perception and multi-agent SLAM.

		\begin{table}[]	
			\centering
			\caption{Dataset Details}
			\label{tab:dataset}			
			\begin{tabular}{@{}llll@{}}
				\toprule
				\textbf{KITTI Odometry}            & \textbf{Train} & \textbf{Validation} & \textbf{Test} \\ \midrule
				\# sequences                       & 5              & 2                   & 2    \\
				\# samples (pairs of point clouds) & 1358           & 180                 & 555 \\ 	\midrule		
				\textbf{CODD}                      & \textbf{Train} & \textbf{Validation} & \textbf{Test} \\ \midrule
				\# exclusive maps                  & 6              & 1                   & 1    \\
				\# sequences                       & 78             & 14                  & 16   \\
				\# samples (pairs of point clouds) & 6129           & 1339                & 1315 \\ \bottomrule	
			\end{tabular}
		\end{table}
	
	\subsection{Evaluation Metrics}
	\label{sec:exp:metrics}
		Following previous studies \cite{choy2019fully,choy2020deep}, the registration performance is evaluated in terms of the translation and rotation errors given by
		\begin{align}
			\label{eq:metrics}
			\text{TE} &= \left\|\hat{t} - t^g\right\|_2, \\
			\text{RE} &= \arccos \frac{\Tr(\hat{R}^\T R^g) - 1}{2},
		\end{align}
		where $R^g,t^g$ denotes the ground-truth rotation matrix and translation vector, respectively.
		These metrics are reported considering their mean value over all dataset samples, denoted as Mean Translation Error (MTE) and Mean Rotation Error (MRE), respectively.
		We also consider the recall rate, measured as the ratio of successful registrations to the total number of samples, where the success criteria is $\text{TE} < 0.6$m and $\text{RE} < 5$deg following \cite{choy2020deep}.
		The runtime performance is evaluated as the average inference time for the registration of a pair of point clouds disregarding the data loading time.
	
	\subsection{Implementation Details}
	\label{sec:exp:implementation}
		Our proposed method is implemented using PyTorch \cite{paszke2019pytorch}, PyTorch Geometric \cite{feyPytorchGeometric}, the CUDA implementation of SA and FP layers from \cite{qi2017pointnetpp} and the Open3D \cite{Zhou2018} Procrustes RANSAC implementation.
		The model is trained independently for each dataset using the Adam \cite{kingma2014adam} optimiser with a learning rate of $0.1$, $\epsilon=10^{-4}$, $\beta_1=0.9, \beta_2=0.999$ and a batch size of 6 (pairs of point clouds).
		The model is trained for twenty epochs and the learning rate is reduced in half after every five epochs.
		For evaluation, the model with the lowest validation loss is selected.
		The temperature hyper-parameter, described in Section \ref{sec:soft-corr}, is set to $T=10^{-2}$, and the loss scaling hyper-parameter is set to $\lambda=10$.
		During inference, the RANSAC inlier threshold, $\kappa$, is set to 0.5m, and the maximum number of RANSAC iterations, denoted by $H$, is computed to achieve $0.999$ confidence in the selected hypothesis within a limit of $10^5$ iterations.
		The point clouds from both datasets are downsampled using voxel sizes of 0.3m following previous methods \cite{choy2019fully,choy2020deep}.
		The overlap ratio distance threshold, $\gamma$, is set to 0.3m, following the down-sampling voxel size. 
		The experiments are carried out on a Xeon ES-1630 CPU and Quadro M4000 GPU with 8 GB of memory.
		For fair comparison, all baselines are also evaluated on the same hardware.
		The official implementation and pre-trained models (30cm voxel) are used for the evaluation of \cite{choy2019fully,choy2020deep} in the KITTI dataset; likewise, the official TEASER \cite{Yang20tro-teaser} implementation is adopted; and the Open3D \cite{Zhou2018} implementation of FPFH, RANSAC and ICP is used for the evaluation of the respective methods.
		
	\subsection{Performance on the KITTI dataset}
	\label{sec:exp:results-kitti}
		The evaluation results, presented in Table \ref{tab:results-kitti}, show that our proposed method achieves competitive registration errors compared to other methods at a significantly lower inference time -- more than five times faster than the fastest baseline.
		While the proposed method has a marginal increase in mean translation error compared to baseline methods, it achieves on-par recall rate relative to baseline methods.
		The mean translation and rotation errors of our method can be further reduced using ICP for refinement (Ours + ICP), at the cost of increased inference time.
		
		\begin{table}[]
			\centering
			\caption{Evaluation Results on KITTI test set}
			\label{tab:results-kitti}
			\resizebox{\linewidth}{!}{%
			\begin{tabular}{@{}lllll@{}}
				\toprule
				\textbf{Model}                      & \textbf{MTE [cm]} & \textbf{MRE [deg]} & \textbf{Recall} & \textbf{Time/sample [s]} \\ \midrule
				FPFH TEASER \cite{Yang20tro-teaser} & 18.8              &  0.76              &  0.984                &  2.26                  \\				
				FCGF RANSAC \cite{choy2019fully}    & 23.7              &  0.034             &  0.975                &  26.29                 \\
				DGR  \cite{choy2020deep}            & 15.6              &  1.43              &  0.982                &  7.60                  \\ \midrule
				Ours                                & 26.1              &  0.74              &  0.949                &  0.41                  \\
				Ours + ICP                          & 8.2               &  0.23              &  0.985                &  3.68                  \\ \bottomrule
			\end{tabular}
			}
		\end{table}
	
	\subsection{Performance on the CODD dataset}
	\label{sec:exp:results-codd}
		We aim to evaluate the performance of the proposed method and baselines on challenging low-overlapping pairs of point clouds.
		For a fair comparison, learning-based methods \cite{choy2019fully,choy2020deep} are also trained on the CODD dataset.
		Table \ref{tab:results} presents the results on the CODD test set, aggregated by the overlap ratio between point clouds in four progressively larger intervals -- the last interval contains all samples.
		Traditional methods are not resilient to low overlapping points clouds, as the registration error increases significantly when considering lower overlap ratios, as shown in the first three rows of Table \ref{tab:results}.
		In contrast, the learning-based baselines are reasonably robust to low overlapping point clouds and achieve high recall rates on all intervals.
		However, the latter methods demand substantial running times due to their complex encoders and the filtering of a high number of putative correspondences.
		In contrast, our proposed method achieves similar or better recall rates to the learning-based baselines with more than 35 times faster inference times.
		This is achieved by our efficient encoder design which outputs a small number of correspondences, which in turn reduces the RANSAC inference time.
		Our efficient encoder strategy comes at the cost of a slight increase of the MTE and MRE metrics, as compared to DGR \cite{choy2020deep}.
		To mitigate this, we apply ICP refinement to our model's output (Ours + ICP), which allows achieving similar MTE and MRE for highly overlapping point clouds and outperforming all baselines on low-overlapping point clouds.
		Although the ICP refinement comes with an additional computational cost, we still achieve a nine-fold speed-up compared to learning-based baselines.
		Qualitative results are presented in Figure \ref{fig:qualitative}.

\begin{figure*}
			\centering
			\input{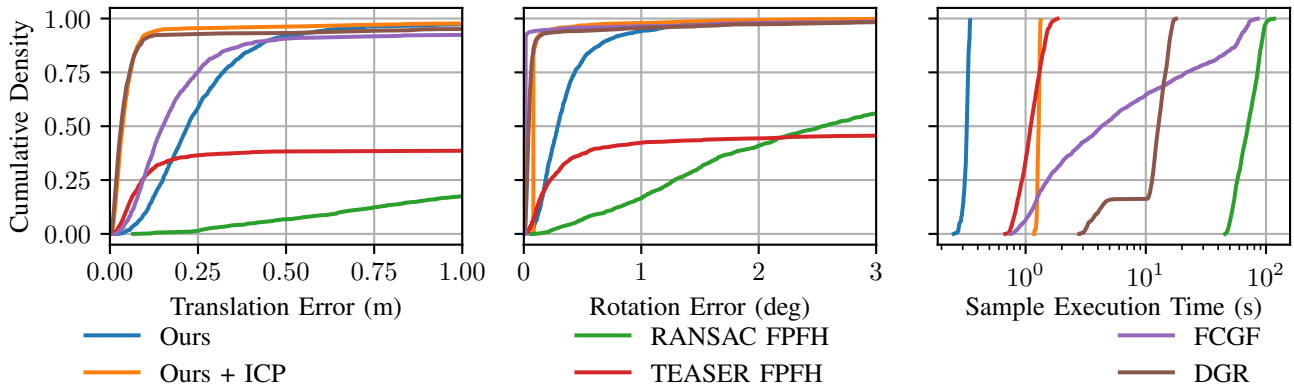}
			\caption{ECDF of the Translation Error, Rotation Error and Sample Execution Time for different methods on the CODD test set.}
			\label{fig:plot:results-ecdf}
		\end{figure*}

		Figure \ref{fig:plot:results-ecdf} shows the Empirical Cumulative Density Function (ECDF) of the translation error, rotation errors, and inference time for different methods.
		The distributions indicate that the proposed method with ICP refinement has the best translation error across samples, closely matched by DGR \cite{choy2020deep}, however with one order of magnitude smaller inference time.
		The inference time distributions show that the proposed method is the fastest among baselines, with an inference time of 320ms on average, with negligible standard deviation (17ms).
		The proposed method can operate in real-time considering data input frequencies up to 3Hz.
		
		\begin{table*}[]
			\centering
			\caption{Evaluation Results on CODD test set}
			\label{tab:results}
			\resizebox{\linewidth}{!}{%
			\begin{tabular}{@{}lllllllllllllll@{}}
			\cmidrule(l){2-15}
			\textbf{} &
			\multicolumn{3}{c}{$\text{Overlap Ratio} > 0.6$} &
			\multicolumn{3}{c}{$\text{Overlap Ratio} > 0.5$} &
			\multicolumn{3}{c}{$\text{Overlap Ratio} > 0.4$} &
			\multicolumn{3}{c}{$\text{Overlap Ratio} > 0$} &
			\multicolumn{2}{c}{\textbf{Time/sample [s]}} \\ \cmidrule(r){1-1} \cmidrule(r){2-4} \cmidrule(r){5-7} \cmidrule(r){8-10} \cmidrule(r){11-13} \cmidrule(r){14-15}
			\textbf{Method} &
			\multicolumn{1}{c}{\textbf{MTE [m]}} &
			\multicolumn{1}{c}{\textbf{MRE [deg]}} &
			\multicolumn{1}{c}{\textbf{Recall}} &
			\multicolumn{1}{c}{\textbf{MTE [m]}} &
			\multicolumn{1}{c}{\textbf{MRE [deg]}} &
			\multicolumn{1}{c}{\textbf{Recall}} &
			\multicolumn{1}{c}{\textbf{MTE [m]}} &
			\multicolumn{1}{c}{\textbf{MRE [deg]}} &
			\multicolumn{1}{c}{\textbf{Recall}} &			
			\multicolumn{1}{c}{\textbf{MTE [m]}} &
			\multicolumn{1}{c}{\textbf{MRE [deg]}} &
			\multicolumn{1}{c}{\textbf{Recall}} &			
			\multicolumn{1}{c}{\textbf{Mean}} &
			\multicolumn{1}{c}{\textbf{Std}} \\ \cmidrule(r){1-1} \cmidrule(r){2-4} \cmidrule(r){5-7} \cmidrule(r){8-10} \cmidrule(r){11-13} \cmidrule(r){14-15}
			ICP \cite{arun1987least}           & 1.69 & 36.11 & 0.67 & 4.81 & 68.67 & 0.38 & 9.11  & 66.42 & 0.17 & 16.14 & 74.84 & 0.07 & 0.38  & 0.048 \\
			RANSAC FPFH \cite{rusu2009fpfh}    & 1.59 & 1.42  & 0.47 & 1.25 & 1.58  & 0.28 & 2.64  & 2.42  & 0.18 & 9.55  & 9.49  & 0.09 & 71.69 & 15.37 \\
			TEASER FPFH \cite{Yang20tro-teaser}& 0.04 & 0.10  & 1.00 & 1.61 & 23.33 & 0.86 & 4.29  & 36.42 & 0.69 & 12.87 & 69.74 & 0.39 & 1.13  & 0.24  \\ \midrule
			RANSAC FCGF \cite{choy2019fully}   & 0.09 & 0.01  & 1.00 & 0.10 & 0.01  & 1.00 & 0.12  & 0.01  & 1.00 & 1.70  & 0.11  & 0.91 & 16.5  & 22.4  \\
			DGR \cite{choy2020deep}            & 0.02 & 0.07  & 1.00 & 0.02 & 0.06  & 1.00 & 0.02  & 0.05  & 1.00 & 0.39  & 1.52  & 0.94 & 11.89 & 3.92  \\ \midrule
			Ours                               & 0.14 & 0.21  & 1.00 & 0.19 & 0.25  & 0.99 & 0.22  & 0.29  & 0.98 & 0.28  & 0.41  & 0.94 & 0.32  & 0.017 \\
			Ours + ICP                         & 0.03 & 0.09  & 1.00 & 0.03 & 0.09  & 0.99 & 0.04  & 0.09  & 0.99 & 0.09  & 0.13  & 0.97 & 1.28  & 0.031 \\ 
			Ours - Att                         & 0.29 & 0.48  & 0.87 & 0.39 & 0.56  & 0.86 & 0.59  & 0.86  & 0.76 & 2.22  & 5.69  & 0.57 & 0.30  & 0.003 \\ \bottomrule
			\end{tabular}
		}
		\end{table*}

	\subsection{Ablation Study}
	\label{sec:exp:results-ablation}
		We assess the impact of the proposed attention network into the registration performance, measured in terms of translation and rotation errors.
		This is achieved by removing the graph attention module, retraining the model and evaluating its performance on the CODD test set.
		The results, indicated in Table \ref{tab:results} ``Ours - Att'', show that the graph-attention network plays a key role in improving the accuracy of the correspondences, resulting in lower translation and rotation errors.
		The benefits of the graph attention network is most significant for low-overlapping point clouds, as indicated by the last range group in Table \ref{tab:results}, where the removal of the attention results in a 40\% reduction of the registration recall and a significant increase in the mean translation and rotation errors.

\section{\uppercase{Conclusion}}
\label{sec:conclusion}
	We proposed a novel point cloud registration method focusing on fast inference of partially overlapping lidar point clouds. 
	The performance evaluation on the KITTI and CODD datasets indicates that the proposed model can operate with a latency lower than 410ms and 320ms, respectively.
	The results show that the proposed model outperform baseline methods in terms of rotation and translation errors for pairs of point clouds with low overlap.
	Furthermore, we show that the proposed graph attention module plays a key role in improving the quality of the correspondences in low overlapping point clouds, which results in higher registration performance.

	\bibliographystyle{IEEEtran}
	\bibliography{root}
	
\end{document}